\documentclass[11pt]{article}

\usepackage[]{acl}

\usepackage{times}
\usepackage{latexsym}
\usepackage[T1]{fontenc}
\usepackage[utf8]{inputenc}
\usepackage{microtype}
\usepackage{graphicx}
\usepackage{amsmath}
\usepackage{amssymb}
\usepackage{booktabs}
\usepackage{array}
\usepackage{tabularx}
\usepackage{colortbl}
\usepackage{xcolor}
\providecommand{\multirow}[3]{#3}
\newcommand{\datasetcell}[1]{\raisebox{-3.2\normalbaselineskip}[0pt][0pt]{#1}}

\title{Event Ontology Expansion via LLM-Based Conceptualization}

\author{
    Weicheng Ren,
    \setcounter{footnote}{1}
    Zixuan Li\thanks{Corresponding authors.},
    Long Bai,
    Xiaolong Jin$^{\dagger}$,
    Jiafeng Guo,
    Xueqi Cheng
     \\
    \textsuperscript{1} State Key Laboratory of AI Safety \\
    \textsuperscript{2} Institute of Computing Technology, Chinese Academy of Sciences \\
    \textsuperscript{3}School of Computer Science and Technology, University of Chinese Academy of Sciences \\
    \small
    \texttt{\{renweicheng21b, lizixuan, bailong, jinxiaolong, guojiafeng, cxq\}@ict.ac.cn}\\
}

\begin{document}
\maketitle

\begin{abstract}
Event ontology expansion aims to discover emerging event types from data and extend them to appropriate positions in the existing event ontology..
Existing methods typically cluster contextualized trigger representations and attach induced clusters to the ontology based on instance-level similarity.
However, ontology expansion requires concept-level semantics that characterize event types, whereas contextualized trigger representations often conflate these semantics with surface contextual variation, leading to unstable clustering and unreliable hierarchy expansion.
To address this issue, we propose ConceptE, a conceptualization-enhanced framework for event ontology expansion.
ConceptE first derives concept-level semantics by prompting an LLM with the sentence and event trigger, producing a concise concept name and a natural-language description.
It then jointly encodes these semantics with trigger information to build concept-enhanced representations aligned with ontology-level reasoning.
This representation design supports more coherent event clustering, more reliable hierarchy expansion, and ontology-consistent type naming.
Experiments on ACE, ERE, and MAVEN demonstrate that ConceptE consistently outperforms state-of-the-art approaches across all subtasks of event ontology expansion.
In particular, it achieves improvements of up to 12.37\% in BCubed-F1 for event clustering and 6.48\% in Taxo\_F1 for hierarchy expansion, demonstrating the effectiveness of the proposed ConceptE method. \footnote{Code available at: \url{https://github.com/ICT-KCG/ConceptE}}
\end{abstract}

\section{Introduction}
\label{sec:intro}
Event extraction is a core task in natural language processing, with broad applications in information retrieval, knowledge graph construction, and recommendation~\cite{goran-etal-ir, liu-etal-event-recommandation, wu-etal-event-graph}.
Most existing event extraction systems operate under a predefined event ontology, where event types and their hierarchical relations are manually specified in advance.
However, such fixed ontologies are difficult to maintain as new data continuously introduces novel event types.
This motivates \emph{event ontology expansion}, which aims to automatically discover new event types from data and integrate them into appropriate positions within an existing ontology hierarchy.

Event ontology expansion can be viewed as a pipeline with three coupled subtasks: event clustering, hierarchy expansion, and type naming.
First, unlabeled event mentions are grouped into clusters, where each cluster corresponds to a candidate emerging event type.
Second, the discovered event types are attached to appropriate positions in the existing ontology hierarchy.
Finally, each newly inserted type is assigned an ontology-consistent name.
This formulation follows recent event ontology completion work~\cite{cao-etal-2023-event} and provides the basis for analyzing the limitations of existing instance-centric methods.


\begin{figure}[t]
  \centering
  \includegraphics[width=1.0\linewidth]{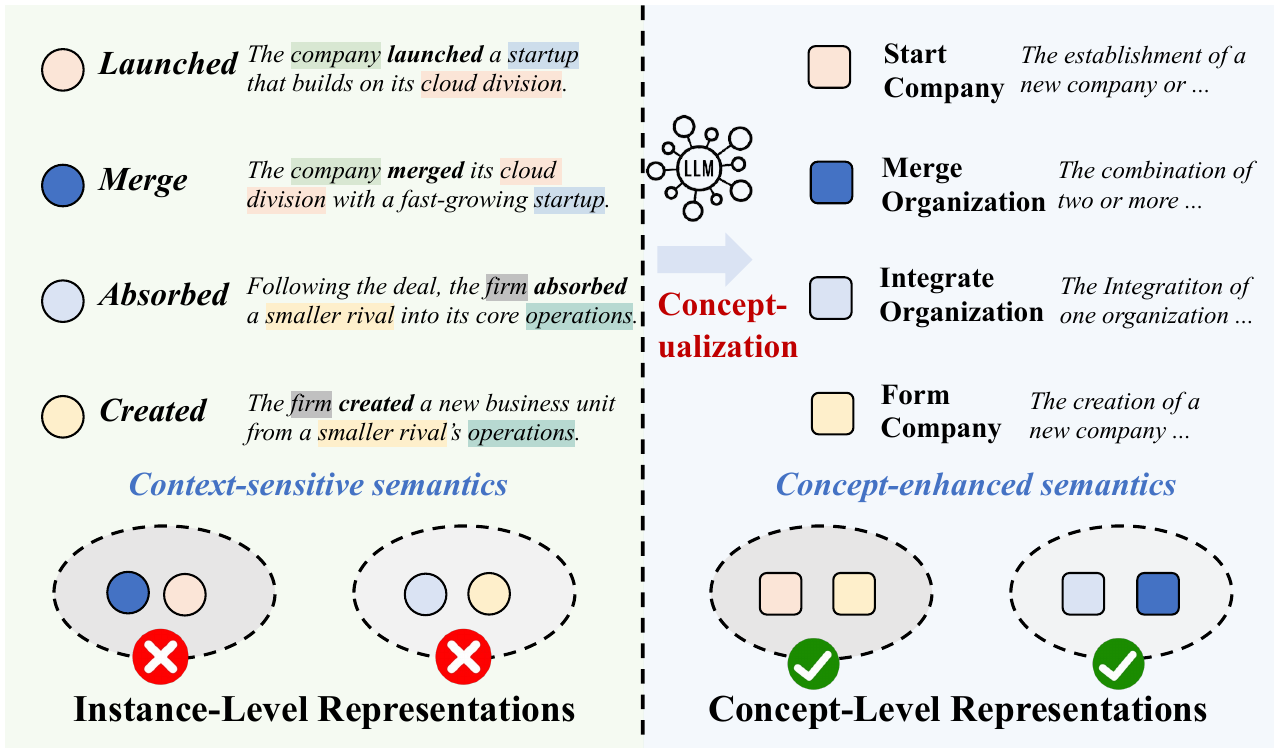}
  \caption{Illustration of the impact of contextual variation on instance-level representations and the stabilizing effect of concept-enhanced representations.}
  \label{fig:motivation}
\end{figure}

Despite recent progress, existing approaches, including HALTON~\cite{cao-etal-2023-event}, remain largely instance-centric.
These methods encode event instances with contextualized trigger representations and use the resulting instance-level similarity for both clustering and hierarchy expansion.
As illustrated in Figure~\ref{fig:motivation} (left), this design is vulnerable to surface contextual variation.
Triggers appearing in similar organization-centered contexts (e.g., \emph{``launched''} and \emph{``merged''} with company-related words) may be grouped together although they express different event types, while semantically related triggers (e.g., \emph{``absorbed''} and \emph{``created''}) may be separated by lexical or contextual differences.

For clustering, this sensitivity can blur the boundaries between event types.
Contextualized trigger representations capture useful trigger cues, but they can overemphasize surrounding entities, lexical cues, and syntactic patterns within the sentence.
As a result, clustering may reflect surface similarity more than the concept-level semantics that characterize event types.

For hierarchy expansion, the same reliance on instance-level similarity makes it difficult to identify the correct parent type.
Attaching a discovered type requires assessing whether a candidate parent represents a broader event category than the new type, rather than selecting a node by embedding similarity alone.
Instance-level similarity may assign high scores to semantically related but structurally incorrect nodes, such as siblings or ancestors.
This motivates a hierarchy expansion module that models directed parent-child attachment instead of undirected similarity matching.

To address these limitations, we use LLMs to derive concept-level semantics for event triggers.
Given a sentence and its event trigger, an LLM generates a concise concept name and a natural-language description that summarize the event type expressed by the trigger.
The derived concept-level semantics are encoded together with the original trigger information to build concept-enhanced representations.
As illustrated in Figure~\ref{fig:motivation} (right), these representations group context-diverse trigger occurrences by shared event-type meaning (e.g., \emph{``Start Company''}, \emph{``Merge Organization''}), enabling more consistent clustering and, in turn, more reliable hierarchy expansion.

We propose ConceptE, a conceptualization-enhanced framework for event ontology expansion.
ConceptE first derives concept-level semantics with LLMs and encodes them together with the original trigger information.
With these concept-enhanced representations, ConceptE clusters event mentions into candidate event types while reducing the effect of surface contextual variation across sentences.

For hierarchy expansion, ConceptE further uses concept-enhanced representations to align newly discovered event types with event types already organized in the existing ontology.
Since hierarchy insertion requires identifying a valid parent type rather than merely finding a similar type, ConceptE models parent-child attachment as a directed relation.
This design places new event types into structurally appropriate ontology positions by using concept-level semantics and directed attachment, rather than symmetric instance-level similarity alone.
The derived concepts also provide semantic signals for ontology-consistent naming.

We evaluate ConceptE on ACE, ERE, and MAVEN, three widely used benchmarks.
Experimental results demonstrate that ConceptE consistently outperforms state-of-the-art approaches across subtasks of event ontology expansion.
In particular, it achieves improvements of up to 12.37\% in BCubed-F1 for event clustering and 6.48\% in Taxo\_F1 for hierarchy expansion, validating the effectiveness of conceptualization for robust event ontology expansion.

\begin{itemize}
  \item We propose ConceptE, a conceptualization-enhanced framework that derives concept-level semantics for event triggers and builds concept-enhanced representations for more stable clustering and hierarchy expansion.
  \item We design a hierarchy expansion module that composes concept-enhanced representations into ontology node representations and learns a directional parent-child linker for directed attachment.
  \item We conduct extensive experiments on ACE, ERE, and MAVEN, demonstrating that ConceptE consistently outperforms strong baselines in event clustering, hierarchy expansion, and event type naming.
\end{itemize}

\section{Task Definition}
\label{sec:task}

\begin{figure*}[htbp]
  \centering
  \includegraphics[width=1.0\linewidth]{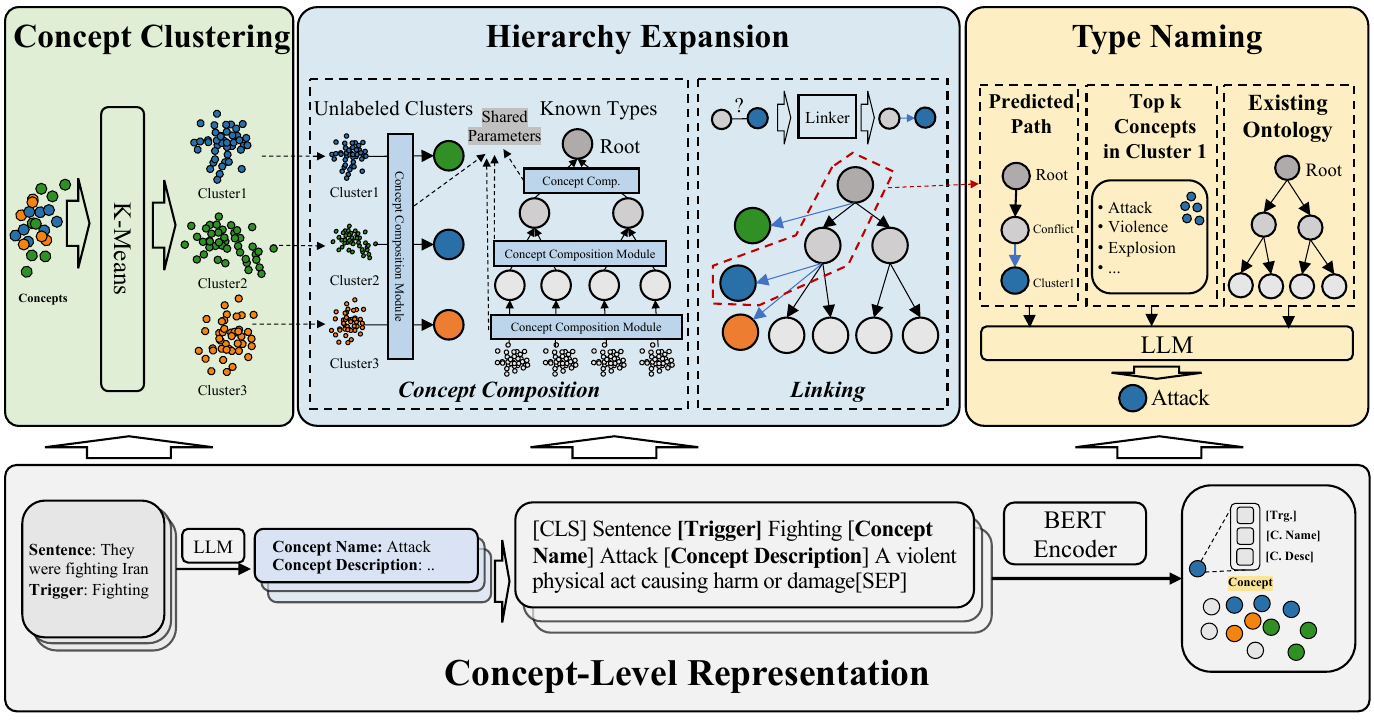}
  \caption{Overall framework of ConceptE. 
  }
  \label{fig:model}
\end{figure*}

Following HALTON~\cite{cao-etal-2023-event}, event ontology expansion takes an existing event ontology $\mathcal{T}_{\text{known}}$, labeled instances of known types $\mathcal{D}_{\text{l}}=\{(x_i,y_i)\}$, and unlabeled instances $\mathcal{D}_{\text{u}}=\{x_j\}$ that may contain unknown event types. The goal is to produce an expanded ontology $\mathcal{T}_{\text{expanded}}$ by discovering novel event types from $\mathcal{D}_{\text{u}}$, attaching them to appropriate positions in $\mathcal{T}_{\text{known}}$, and generating ontology-consistent type names. Thus, the task jointly involves event clustering, hierarchy expansion, and type naming.

\section{Method}
\label{sec:method}


We propose \textbf{ConceptE}, a conceptualization-enhanced framework that derives concept-level semantics for event ontology expansion.
ConceptE first derives concept-level semantics for event triggers with LLMs, and then applies concept-enhanced representations to three subtasks: (i) \emph{Concept Clustering}, (ii) \emph{Hierarchy Expansion}, and (iii) \emph{Type Naming}.
Figure~\ref{fig:model} illustrates the overall architecture.


\subsection{Concept-Level Representation}

Contextualized trigger representations provide important trigger-level information, but they are sensitive to surrounding entities and syntactic patterns within the sentence, which can scatter instances of the same event type and mislead downstream hierarchy expansion.
ConceptE enriches these representations by leveraging an LLM to infer the concept-level semantics expressed by each event trigger.
Specifically, the prompt provides only the sentence and the event trigger and instructs the LLM to summarize the trigger occurrence into a concise \emph{concept name} and a natural-language \emph{concept description}.
The concept name serves as a compact candidate type phrase, while the description explains the event semantics grounded in the sentence containing the trigger.
The full prompt and granularity constraints are provided in Appendix~\ref{app:concept_details}.

Based on the generated concepts, we concatenate the original sentence and trigger information with the generated concept-level semantics:
\begin{quote}\small
[CLS] sentence
[Trigger] trigger
[Concept Name] short descriptor
[Concept Description] semantic abstraction
[SEP]
\end{quote}

This input retains the original sentence and trigger as direct textual information while representing concept-level semantics through the generated concept name and description.
We encode it using a pretrained contextual encoder (e.g., BERT~\cite{devlin-etal-2019-bert}), where the sentence, trigger, concept name, and concept description are jointly modeled through self-attention.
Let $\mathbf{h}_{\text{t}}$, $\mathbf{h}_{\text{cn}}$, and $\mathbf{h}_{\text{cd}}$ denote the pooled representations of the trigger span, the generated concept-name span, and the generated concept-description span, respectively.
We concatenate these representations and project them back to the original embedding space:
\begin{equation}
\mathbf{h} = \mathbf{W}_{\text{fuse}}
\left[
\mathbf{h}_{\text{t}} \, \| \,
\mathbf{h}_{\text{cn}} \, \| \,
\mathbf{h}_{\text{cd}}
\right] + \mathbf{b}_{\text{fuse}},
\end{equation}
where $[\cdot \| \cdot]$ denotes vector concatenation, and $\mathbf{W}_{\text{fuse}} \in \mathbb{R}^{d \times 3d}$ is a trainable projection matrix.

The fused representation $\mathbf{h}$ therefore combines trigger-level information with reusable concept-level semantics, providing a more suitable representation for ontology-level reasoning.

\subsection{Concept Clustering}
\label{sec:cluster_obj}

To discover coherent candidate event types under contextual variation, we perform event clustering based on concept-enhanced representations.
Following the neighborhood contrastive clustering framework in HALTON~\cite{cao-etal-2023-event}, we jointly leverage labeled and unlabeled instances to learn discriminative clustering objectives over these representations.

Let $\mathbf{h}_i$ denote the concept-enhanced representation of instance $x_i$ obtained from the previous module. Given labeled instances $\mathcal{D}_l=\{(x_i^l, y_i^l)\}$, we apply a standard cross-entropy loss:
\begin{equation}
\mathcal{L}_{\mathrm{ce}} = - \frac{1}{|\mathcal{D}_l|} \sum_i y_i^l \log \mathrm{softmax}(\mathbf{h}_i^l).
\end{equation}

For unlabeled instances $\mathcal{D}_u=\{x_i^u\}$, we update pseudo cluster assignments at the beginning of each epoch.
Specifically, we encode all unlabeled instances into the clustering space and run $k$-means with $k=|\mathcal{Y}_{\text{new}}|$.
The resulting pseudo label $\hat{y}_i$ is stored with each unlabeled instance and used to construct pairwise supervision: $q_{ij}=1$ if $\hat{y}_i=\hat{y}_j$, and $q_{ij}=0$ otherwise.
Following HALTON~\cite{cao-etal-2023-event}, we adopt a hinge-style contrastive objective with pairwise supervision to encourage instances from the same pseudo cluster to produce similar predictions:
\begin{equation}
\mathcal{L}_{\mathrm{hng}} = \sum_{i,j} \Big( q_{ij} d_{ij} + (1-q_{ij}) \max(0, \alpha - d_{ij}) \Big),
\end{equation}
where $d_{ij}$ is the bidirectional KL divergence between the predicted cluster distributions of $x_i$ and $x_j$, and $\alpha$ is the margin for pairs from different pseudo clusters.

To further enhance representation compactness, we introduce neighborhood contrastive losses for both unlabeled
and labeled data.
For unlabeled instances, positives in Eq.~\ref{eq:ncu} are determined by nearest-neighbor mining rather than gold labels.
We maintain a memory bank of current instance representations and retrieve the top-$K$ nearest neighbors for each unlabeled instance.
Within a mini-batch, an unlabeled instance $x_j$ is treated as a positive of $x_i$ if the global index of $x_j$ belongs to this top-$K$ neighbor set $\mathcal{N}_i$; the instance itself is also included to form the diagonal positive.
In training, we additionally sample one neighbor from $\mathcal{N}_i$ as a second view for supervised contrastive learning.
The unlabeled neighborhood contrastive loss is defined as:
\begin{equation}
\mathcal{L}_{\mathrm{ncu}}
= - \sum_i \sum_{j \in \mathcal{N}_i}
\log
\frac{\exp(r^u_{ij}/\tau)}
{\sum_{k \neq i}\exp(r^u_{ik}/\tau)}.
\label{eq:ncu}
\end{equation}
\begin{equation}
r^u_{ij} = \mathrm{sim}(\mathbf{h}_i^u,\mathbf{h}_j^u).
\end{equation}

For labeled instances, positives are all in-batch instances sharing the same gold type:
\begin{equation}
\mathcal{L}_{\mathrm{ncl}}
= - \sum_i \sum_{j \in \mathcal{N}_i^l}
\log
\frac{\exp(r^l_{ij}/\tau)}
{\sum_{k \neq i}\exp(r^l_{ik}/\tau)}.
\end{equation}
\begin{equation}
r^l_{ij} = \mathrm{sim}(\mathbf{h}_i^l,\mathbf{h}_j^l).
\end{equation}

These objectives jointly encourage intra-type compactness and inter-type separability under both supervised and unsupervised settings.

\begin{equation}
\mathcal{L}_{\mathrm{cluster}} 
= \mathcal{L}_{\mathrm{ce}} + \mathcal{L}_{\mathrm{hng}} + \mathcal{L}_{\mathrm{ncu}} + \mathcal{L}_{\mathrm{ncl}}.
\end{equation}

\subsection{Hierarchy Expansion}
\label{sec:taxonomy}

After event clustering, each cluster corresponds to a latent event type that must be integrated into the existing ontology.
A naive approach attaches clusters based on instance-level similarity.
However, such similarity does not explicitly represent semantic scope or attachment direction, and therefore often results in structurally inconsistent expansions.
Hierarchy expansion requires deciding whether an existing ontology node can subsume a newly discovered event type.
To address this requirement, we explicitly model ontology nodes as semantic units and learn directed parent-child relations based on concept-enhanced representations, allowing newly induced event types to be attached to structurally and semantically appropriate positions.

\subsubsection{Concept Composition}
\label{sec:comp}

To enable hierarchy-aware linking, we introduce a Concept Composition module that constructs semantic representations for ontology nodes in a bottom-up manner.
We model the event ontology as a rooted tree, where leaf nodes correspond to known event types or newly discovered clusters and are initialized from their associated instance representations, while internal nodes recursively compose semantics from their children.
Newly discovered clusters and known event types are treated uniformly as leaf nodes, and the same Concept Composition module with shared parameters is applied across all hierarchy levels to ensure consistent semantic composition.




Let $\mathcal{C}(v)$ denote the children of node $v$.
We implement composition with attention-weighted pooling over child embeddings and combine it with residual mean pooling:
\begin{equation}
\tilde{\mathbf{h}}_v = \sum_{u\in\mathcal{C}(v)} \alpha_u\mathbf{h}_u,
\end{equation}
\begin{equation}
\mathbf{h}_v = \mathrm{LayerNorm}\big(
\mathrm{Proj}(\tilde{\mathbf{h}}_v)
+ \mathrm{Mean}(\{\mathbf{h}_u\})\big),
\end{equation}
where $\alpha_u$ is a learned attention weight normalized over children and $\mathrm{Proj}(\cdot)$ is a linear projection layer.
Through this recursive composition process, explicit semantic representations are obtained for all ontology nodes.

\subsubsection{Linking}
\label{sec:linker}

To model the asymmetric semantic relation that a child event type is subsumed by its parent type, we design a directional parent-child linker.
Given normalized child and parent embeddings $\mathbf{u}$ and $\mathbf{v}$, we construct interaction features:
\begin{equation}
\mathbf{z}=[\mathbf{u};\mathbf{v};\mathbf{u}\odot\mathbf{v};|\mathbf{u}-\mathbf{v}|],
\end{equation}
where $[\cdot;\cdot]$ denotes concatenation and $\odot$ denotes elementwise product.
A lightweight MLP then produces a directed compatibility score $s(c\rightarrow p)$.

For each newly discovered cluster represented by its embedding $\mathbf{h}_{\text{new}}$, we perform greedy top-down inference starting from the ontology root.
At each step, we score the new type against each child of the current node as a candidate parent, descend to the best-scoring child if it improves over the current node, and stop otherwise.
The final node determines the attachment position of the new event type under greedy top-down inference.

\subsubsection{Training Objectives}
\label{sec:taxo_obj}

We train the hierarchy module with two objectives.
The composition loss aligns composed node embeddings with gold type-name embeddings, while the linker loss uses existing parent-child edges as positives and non-ancestor nodes as negatives.
The overall objective is:
\begin{equation}
\mathcal{L}_{\mathrm{taxo}} = \lambda_{\mathrm{comp}}\,\mathcal{L}_{\mathrm{comp}} + \lambda_{\mathrm{link}}\,\mathcal{L}_{\mathrm{link}},
\end{equation}
where $\mathcal{L}_{\mathrm{comp}}$ is a cosine alignment loss and $\mathcal{L}_{\mathrm{link}}$ is a weighted binary cross-entropy loss.
\subsection{Type Naming}
\label{sec:naming}

After clustering and hierarchy expansion, each newly discovered cluster corresponds to a latent event type that requires an ontology-consistent name.
The goal of this stage is to convert the latent cluster into a human-readable event type label that matches the granularity and naming style of the existing ontology.
This is challenging because a cluster may contain diverse triggers and contexts, and a single representative instance may not fully capture the cluster semantics.

We therefore perform ontology-aware type naming at the cluster level.
For each discovered cluster, we aggregate high-frequency concept names as semantic anchors and combine them with the predicted ontology path as structural context.
The concept set summarizes what the cluster expresses, while the parent path constrains the naming space to an appropriate ontology region.
An LLM then generates a concise type name conditioned on both sources of information.
This design encourages names that are semantically faithful to the cluster and structurally consistent with the ontology.
Detailed prompt fields and examples are provided in Appendix~\ref{app:type_naming_details}.

\section{Experimental Setup}
\label{sec:exp}

\begin{table*}[htbp]
\centering
\caption{Event clustering performance.}
\label{tab:cluster_performance}
\resizebox{0.8\textwidth}{!}{
\begin{tabular}{l|l|cccc}
\toprule
Datasets & Methods & ARI (\%) & NMI (\%) & Accuracy (\%) & BCubed-F1 (\%) \\
\midrule
\multirow{5}{*}{ACE} & SS-VQ-VAE & 8.53 & 33.81 & 29.95 & 27.60 \\
& ETYPECLUS & 26.17 & 53.91 & 40.70 & 38.69 \\
& TABS & 59.18 & 79.36 & 71.42 & 69.44 \\
& HALTON & 67.41 & 84.29 & 77.26 & 75.06 \\
\rowcolor[gray]{0.9} & \textbf{ConceptE (Ours)} & \textbf{76.27 ($\uparrow$ 8.86)} & \textbf{89.87 ($\uparrow$ 5.58)} & \textbf{84.58 ($\uparrow$ 7.32)} & \textbf{83.25 ($\uparrow$ 8.19)} \\
\midrule\midrule
\multirow{5}{*}{ERE} & SS-VQ-VAE & 13.46 & 40.45 & 29.96 & 26.69 \\
& ETYPECLUS & 15.89 & 46.86 & 34.55 & 29.13 \\
& TABS & 47.22 & 71.26 & 60.24 & 55.82 \\
& HALTON & 56.01 & 78.13 & 67.72 & 64.66 \\
\rowcolor[gray]{0.9} & \textbf{ConceptE (Ours)} & \textbf{65.61 ($\uparrow$ 9.60)} & \textbf{84.02 ($\uparrow$ 5.89)} & \textbf{75.93 ($\uparrow$ 8.21)} & \textbf{73.33 ($\uparrow$ 8.67)} \\
\midrule\midrule
\multirow{5}{*}{MAVEN} & SS-VQ-VAE & 3.06 & 17.57 & 12.29 & 11.14 \\
& ETYPECLUS & 11.27 & 30.79 & 20.82 & 14.73 \\
& TABS & 27.93 & 53.84 & 39.38 & 31.52 \\
& HALTON & 36.03 & 60.34 & 52.70 & 39.35 \\
\rowcolor[gray]{0.9} & \textbf{ConceptE (Ours)} & \textbf{46.99 ($\uparrow$ 10.96)} & \textbf{69.36 ($\uparrow$ 9.02)} & \textbf{62.81 ($\uparrow$ 10.11)} & \textbf{51.72 ($\uparrow$ 12.37)} \\
\bottomrule
\end{tabular}
}
\end{table*}

\subsection{Datasets}
\label{sec:data}

We follow the dataset construction in HALTON~\cite{cao-etal-2023-event} based on ACE~\cite{doddington-etal-2004-automatic}, ERE~\cite{song-etal-2015-light}, and MAVEN~\cite{wang-etal-2020-maven}.
For ACE and ERE, the top 10 most frequent types are treated as known, with the remaining types as unknown.
For MAVEN, the top 20 are known, and the remaining 40 are unknown, to mitigate long-tail effects.
Dataset statistics are provided in Appendix~\ref{app:datasets}.


\subsection{Baselines}
\label{sec:baselines}

We compare ConceptE with representative baselines for the three subtasks.
For event clustering, we include SS-VQ-VAE~\cite{huang-ji-2020-semi}, ETYPECLUS~\cite{shen-etal-2021-corpus}, TABS~\cite{li-etal-2021-document}, and HALTON~\cite{cao-etal-2023-event}, covering semi-supervised, unsupervised, abstraction-enhanced, and recent ontology-expansion settings.
For hierarchy expansion, we compare with clustering baselines combined with greedy expansion (GE), Type\_Similarity, LLMs\_Prompt, and HALTON.
For type naming, we compare against TABS, T5\_Template~\cite{Raffel-etal-2020-exploring}, Trigger\_Sel, HALTON, and Top1\_Concept.
Detailed descriptions are provided in Appendix~\ref{app:baselines}.

\subsection{Evaluation Metrics}
\label{sec:metrics}
\paragraph{Clustering metrics.}
Following HALTON~\cite{cao-etal-2023-event}, we evaluate clustering performance using Adjusted Rand Index (ARI)~\cite{hubert1985comparing}, BCubed-F1~\cite{bagga-baldwin-1998-entity}, Accuracy~\cite{crouse2016implementing}, and NMI~\cite{huang-ji-2020-semi}.

\paragraph{Taxonomy metrics.}
For hierarchy expansion, we adopt the taxonomy metric~\cite{2016evalonto} to evaluate parent-child linking accuracy in the ontology, and report Taxo\_P, Taxo\_R, and Taxo\_F1.

\paragraph{Naming metrics.}
For event type naming, we report ROUGE-L~\cite{lin-2004-rouge} and BERTScore~\cite{bert-score} to measure lexical overlap and semantic similarity between predicted and gold type names.

\subsection{Implementation Details}
We use BERT-base-uncased~\cite{devlin-etal-2019-bert} as the encoder and Qwen3-32B~\cite{Yang2025Qwen3} for conceptualization and type naming.
Full hyperparameters and training details are provided in Appendix~\ref{app:implementation}. We report single-run results due to computational constraints.

\begin{table*}[htbp]
\centering
\caption{Hierarchy expansion performance.}
\label{tab:taxonomy_performance}
\small
\setlength{\tabcolsep}{3.8pt}
\renewcommand{\arraystretch}{1.05}
\resizebox{\textwidth}{!}{%
\begin{tabular}{l|l|ccc|ccc}
\toprule
\textbf{Datasets} & \textbf{Methods} & \multicolumn{3}{c|}{\textbf{Predicted Cluster}} & \multicolumn{3}{c}{\textbf{Golden Cluster}} \\
\cmidrule{3-8}
& & \textbf{Taxo\_P (\%)} & \textbf{Taxo\_R (\%)} & \textbf{Taxo\_F1 (\%)} & \textbf{Taxo\_P (\%)} & \textbf{Taxo\_R (\%)} & \textbf{Taxo\_F1 (\%)} \\
\midrule
\datasetcell{ACE} & SS-VQ-VAE+GE & 9.12 & 10.14 & 9.60 & 9.52 & 13.04 & 11.01 \\
& ETYPECLUS+GE & 30.70 & 23.46 & 26.59 & 34.14 & 33.33 & 33.73 \\
& TABS+GE & 34.31 & 30.43 & 32.25 & 33.33 & 37.68 & 35.37 \\
\cmidrule{2-8}
& Type\_Similarity & 31.79 & 40.58 & 35.65 & 33.33 & 40.37 & 36.51 \\
& LLMs\_Prompt & 34.09 & 34.78 & 34.43 & 42.85 & 43.47 & 43.16 \\
\cmidrule{2-8}
& HALTON &  37.00 & 39.13 & 38.04 & 44.44 & 44.92 & 44.68 \\
\rowcolor[gray]{0.9}\cellcolor{white} & \textbf{ConceptE (Ours)} & \textbf{43.48} & \textbf{43.48} & \textbf{43.48 ($\uparrow$ 5.44)} & \textbf{55.00} & \textbf{47.83} & \textbf{51.16 ($\uparrow$ 6.48)} \\
\midrule\midrule
\datasetcell{ERE} & SS-VQ-VAE+GE & 16.38 & 14.28 & 15.26 & 26.00 & 25.00 & 25.49 \\
& ETYPECLUS+GE & 9.85 & 9.52 & 9.68 & 18.00 & 16.66 & 17.30 \\
& TABS+GE & 23.68 & 17.85 & 20.36 & 26.00 & 25.00 & 25.49 \\
\cmidrule{2-8}
& Type\_Similarity & 20.37 & 21.49 & 20.88 & 22.00 & 21.42 & 21.71 \\
& LLMs\_Prompt & 20.68 & 20.43 & 20.55 & 24.00 & 21.49 & 22.64 \\
\cmidrule{2-8}
&  HALTON & 22.54 &23.60& 23.06 & 26.80 & 25.73 & 26.25 \\
\rowcolor[gray]{0.9}\cellcolor{white} & \textbf{ConceptE (Ours)} & \textbf{28.00} & \textbf{25.00} & \textbf{26.41 ($\uparrow$ 3.35)} & \textbf{32.00} & \textbf{28.57} & \textbf{30.19 ($\uparrow$ 3.94)} \\
\midrule\midrule
\datasetcell{MAVEN} & SS-VQ-VAE+GE & 19.45 & 20.14 & 19.79 & 26.94 & 43.00 & 33.13 \\
& ETYPECLUS+GE & 15.83 & 17.50 & 16.62 & 23.75 & 28.75 & 26.01 \\
& TABS+GE & 27.82 & 32.03 & 29.78 & 27.53 & 40.42 & 32.75 \\
\cmidrule{2-8}
& Type\_Similarity & 22.50 & 27.50 & 24.75 & 27.91 & 32.50 & 30.03 \\
& LLMs\_Prompt & 12.50 & 10.00 & 11.11 & 27.50 & 21.50 & 23.97 \\
\cmidrule{2-8}
&  HALTON & 34.79 & 52.50 & 41.85 & 39.38 & \textbf{59.38} & 47.35\\
\rowcolor[gray]{0.9}\cellcolor{white} & \textbf{ConceptE (Ours)} & \textbf{41.25} & \textbf{55.42} & \textbf{47.30 ($\uparrow$ 5.45)} & \textbf{43.75} & 59.17 & \textbf{50.30 ($\uparrow$ 2.95)}\\
\bottomrule
\end{tabular}
}
\end{table*}

\section{Results and Analysis}
\label{sec:results}

\subsection{Event Clustering Results}
\label{sec:cluster_results}

Table~\ref{tab:cluster_performance} shows that ConceptE consistently outperforms all clustering baselines across ACE, ERE, and MAVEN.
Compared with HALTON, ConceptE improves ARI by $8$--$11\%$ and achieves a $12.37\%$ BCubed-F1 gain on MAVEN, indicating that concept-enhanced representations improve both global cluster alignment and instance-level coherence.
These improvements are consistent across datasets with different ontology sizes and unknown-type ratios, suggesting that ConceptE does not rely on dataset-specific type distributions.
In particular, the gains on MAVEN are important because MAVEN contains more event types and a more complex ontology, making clustering more vulnerable to contextual variation.
These results support the core motivation of ConceptE: modeling concept-level semantics together with trigger information helps reduce within-type variance while preserving discriminative information between types.

\subsection{Hierarchy Expansion Results}
\label{sec:taxo_results}

Table~\ref{tab:taxonomy_performance} reports hierarchy expansion results under Predicted Cluster and Golden Cluster settings.
ConceptE achieves the best Taxo\_F1 across datasets and improves over HALTON by $3.35$--$5.45\%$ with predicted clusters and $2.95$--$6.48\%$ with golden clusters.
The gains under Golden Cluster indicate that ConceptE improves hierarchy modeling beyond the effect of better clustering.
Compared with Type\_Similarity and LLMs\_Prompt, ConceptE benefits from explicitly modeling ontology nodes and directed parent-child relations rather than relying only on embedding similarity or direct prompting.
The Predicted Cluster setting reflects the full pipeline, where hierarchy expansion must tolerate upstream clustering errors.
The consistent improvements in this setting show that concept-enhanced representations provide more reliable cluster semantics for attachment.
Meanwhile, the Golden Cluster setting isolates the hierarchy module, and the remaining gains demonstrate that recursive concept composition and directional linking are useful even when clustering is assumed to be correct.

\subsection{Naming Results}
\label{sec:naming_results}

Table~\ref{tab:naming_results} compares ConceptE with naming baselines using ROUGE-L and BERTScore.
ConceptE consistently outperforms Top1\_Concept and HALTON, showing that multiple representative concepts and predicted ontology paths provide stronger semantic coverage and structural constraints for naming.
Top1\_Concept performs reasonably well because the most frequent concept often captures the dominant semantics of a cluster.
However, relying on a single concept can miss sub-patterns within the cluster or produce names that are too narrow.
ConceptE instead conditions generation on multiple representative concepts and the predicted path, which encourages names that match both the cluster semantics and the naming style of the existing ontology.

\begin{table}[t]
\centering
\small
\caption{Event type naming performance.}
\label{tab:naming_results}
\resizebox{\linewidth}{!}{%
\begin{tabular}{l|l|cc}
\toprule
\textbf{Datasets} & \textbf{Methods} & \textbf{ROUGE-L (\%)} & \textbf{BERTScore (\%)} \\
\midrule
\multirow{4}{*}{ACE} & TABS & 17.49 & 29.40 \\
 & T5\_Template & 18.66 & 35.25 \\
 & Trigger\_Sel & 20.86 & 42.46 \\
\cmidrule{2-4}
 & HALTON & 24.09 & 46.24 \\
 & Top1\_Concept  & 26.21 & 48.72 \\
\rowcolor[gray]{0.9} & ConceptE (Ours)  & \textbf{30.43($\uparrow$ 4.22)} & \textbf{53.33($\uparrow$ 4.61)} \\
\midrule
\midrule
\multirow{4}{*}{ERE} & TABS & 11.90 & 28.03 \\
 & T5\_Template & 13.46 & 32.51 \\
 & Trigger\_Sel & 12.59 & 35.07 \\
\cmidrule{2-4}
 & HALTON & 16.20 & 39.32 \\
 & Top1\_Concept  & 16.43 & 39.87\\
\rowcolor[gray]{0.9} & ConceptE (Ours)  & \textbf{17.66($\uparrow$ 1.23)} & \textbf{41.21($\uparrow$ 1.34)} \\
\midrule
\midrule
\multirow{4}{*}{MAVEN} & TABS & 16.02 & 36.24 \\
 & T5\_Template & 24.94 & 38.20 \\
 & Trigger\_Sel & 27.30 & 40.70 \\
\cmidrule{2-4}
 & HALTON & 30.89 & 41.14 \\
 & Top1\_Concept  & 31.12 & 41.53\\
\rowcolor[gray]{0.9} & ConceptE (Ours)  & \textbf{33.36($\uparrow$ 2.24)} & \textbf{43.96($\uparrow$ 2.43)} \\
\bottomrule
\end{tabular}
}
\end{table}

\subsection{Ablation Studies}
\label{sec:ablation}
We conduct ablation studies to assess conceptualization, concept composition, and the parent-child linker.
As shown in Table~\ref{tab:alb_cluster}, removing concept information consistently degrades clustering performance on ACE, with BCubed-F1 dropping from 83.25\% to 74.23\%.
Table~\ref{tab:alb_relation} further shows that conceptualization also benefits hierarchy expansion.
These drops confirm that concepts are not merely auxiliary text for naming, but also improve the learned representation space used by clustering and hierarchy expansion.
Without conceptualization, the model loses explicit concept-level semantics and becomes more susceptible to surface lexical and syntactic variation.

\begin{table}[t]
    \centering
    \small
    \caption{Ablation study by removing Conceptualization in event clustering on the ACE dataset.}
    \label{tab:alb_cluster}
    \resizebox{\linewidth}{!}{%
    \begin{tabular}{lcccc}
        \toprule
        Method & ARI (\%) & NMI (\%) & Accuracy (\%) & BCubed-F1 (\%) \\
        \midrule
        \textbf{ConceptE} & \textbf{76.27} & \textbf{89.87} & \textbf{84.58} & \textbf{83.25} \\
        \quad w/o Concept & 66.20 & 84.08 & 76.79 & 74.23 \\
        \bottomrule
    \end{tabular}
    }
\end{table}

Removing either concept composition or the directional linker also hurts Taxo\_F1, with the largest drop from removing the linker.
This confirms that recursive node composition and asymmetric parent-child modeling are both important for reliable hierarchy expansion.
The sharp degradation without the linker suggests that symmetric similarity is insufficient for hierarchy insertion, where the direction from child to parent matters.
The composition module also contributes because internal ontology nodes need explicit semantic representations rather than simple averages over leaf instances.

\begin{table}[t]
    \centering
    \small
    \caption{Ablation study by removing Conceptualization, Concept Composition, and Linker in Hierarchy Expansion on the ACE dataset.}
    \label{tab:alb_relation}
    \resizebox{\linewidth}{!}{%
    \begin{tabular}{l|c|c}
        \toprule
        \multirow{2}{*}{Methods} & \multicolumn{2}{c}{Taxo\_F1 (\%)} \\ 
                                & Predicted Cluster & Golden Cluster \\ 
        \midrule
        \textbf{ConceptE}               & \textbf{43.48}             & \textbf{51.16}          \\ 
        \quad w/o Concept             & 34.78             & 41.86          \\ 
        \quad w/o Comp.                & 39.13             & 46.51          \\ 
        \quad w/o Linker              & 14.35              & 19.09           \\ 
        \quad w/o Comp.\&Linker        & 36.67             & 43.01          \\ 
        \bottomrule
    \end{tabular}
    }
\end{table}

\subsection{Analysis of Interaction Features in the Linker}

We further analyze the interaction features used by the parent-child linker in Appendix~\ref{app:feature_analysis}.
The results show that combining node identity features with compatibility and mismatch features yields the best hierarchy expansion performance.
This supports the design of the linker, where parent-child attachment requires more than symmetric embedding similarity.

\subsection{Event Clustering Visualization}

We further visualize event representations on the ERE dataset with t-SNE under multiple perplexity values.
As shown in Appendix~\ref{app:visualization}, ConceptE consistently forms more compact and better separated clusters than HALTON, supporting the quantitative improvements in Table~\ref{tab:cluster_performance}.
Although t-SNE provides only qualitative evidence, the trend remains stable across different perplexity values, suggesting that the benefit of conceptualization is not an artifact of a single visualization setting.

\section{Related Work}
\label{sec:related}

Early work relevant to event ontology expansion primarily focused on event type induction and ontology induction, which discover latent event types from unlabeled or weakly labeled corpora~\cite{huang-ji-2020-semi, shen-etal-2021-corpus, edwards-ji-2023-semi, xu-etal-2024-ceo}.
These methods reduce reliance on fixed ontologies, but mainly focus on clustering or constructing ontologies from scratch rather than expanding a given ontology with newly discovered event types.
Taxonomy expansion methods instead insert provided new concepts into an existing taxonomy~\cite{shen-etal-2020-taxoexpan, yu-etal-2020-steam, manzoor-etal-2020-expanding, zeng-etal-2025-codetaxo}.
This assumption is less suitable for event ontologies, where new types often emerge from data.

The most related work, HALTON~\cite{cao-etal-2023-event}, discovers new event types, attaches them to an existing event ontology, and generates type names.
However, HALTON remains instance-centric: both clustering and hierarchy expansion rely on contextualized instance representations, which are sensitive to context variation.
ConceptE differs by deriving concept-level semantics with LLMs and building concept-enhanced representations for event ontology expansion.
\section{Conclusions}
\label{sec:conclusions}

This work proposes ConceptE, a conceptualization-enhanced framework for event ontology expansion.
ConceptE uses LLMs to derive concept-level semantics for event triggers and jointly encodes them with trigger information to build concept-enhanced representations.
By reducing surface contextual variation, ConceptE supports more coherent event clustering and provides reliable semantic units for hierarchy expansion.
For hierarchy expansion, ConceptE further uses recursive concept composition and a directional parent-child linker to attach newly discovered event types to structurally appropriate ontology positions.
Experiments and ablation studies on three benchmarks demonstrate the effectiveness of conceptualization as a semantic bridge between event triggers and ontology structures.


\section*{Limitations}
Despite its effectiveness, ConceptE relies on LLMs for event conceptualization and type naming, which may introduce extra preprocessing cost when scaling to very large corpora.
The quality of generated concepts can also be affected by model bias or occasional generation inconsistency, although our representation learning module mitigates this risk by jointly encoding the original trigger information and the generated concept-level semantics.
Following prior event ontology expansion methods such as HALTON~\cite{cao-etal-2023-event}, our experiments assume that event triggers are given.
This setting enables a controlled comparison with existing methods, but future work can integrate automatic trigger detection to support a more end-to-end expansion pipeline.

\section*{Ethical Considerations}
This work focuses on automatic event ontology expansion from text and does not involve human subjects or personally identifiable information. All experiments are conducted on publicly available benchmark datasets widely used in prior research. The proposed framework organizes event types at an abstract semantic level and does not generate new factual claims about real-world entities.

\bibliography{main_reference}

@String{Computing = "Computing" }

@String{Springer = "Springer-Verlag" }

@ArtifactSoftware{R,
    title = {R: A Language and Environment for Statistical Computing},
    author = {{R Core Team}},
    organization = {R Foundation for Statistical Computing},
    address = {Vienna, Austria},
    year = {2019},
    url = {https://www.R-project.org/},
}

@inproceedings{cao-etal-2023-event,
    title = "Event Ontology Completion with Hierarchical Structure Evolution Networks",
    author = "Cao, Pengfei  and
      Hao, Yupu  and
      Chen, Yubo  and
      Liu, Kang  and
      Xu, Jiexin  and
      Li, Huaijun  and
      Jiang, Xiaojian  and
      Zhao, Jun",
    editor = "Bouamor, Houda  and
      Pino, Juan  and
      Bali, Kalika",
    booktitle = "Proceedings of the 2023 Conference on Empirical Methods in Natural Language Processing",
    month = dec,
    year = "2023",
    address = "Singapore",
    publisher = "Association for Computational Linguistics",
    url = "https://aclanthology.org/2023.emnlp-main.21/",
    doi = "10.18653/v1/2023.emnlp-main.21",
    pages = "306--320"
}

@inproceedings{xu-etal-2024-ceo,
    title = "{CEO}: Corpus-based Open-Domain Event Ontology Induction",
    author = "Xu, Nan  and
      Zhang, Hongming  and
      Chen, Jianshu",
    editor = "Graham, Yvette  and
      Purver, Matthew",
    booktitle = "Findings of the Association for Computational Linguistics: EACL 2024",
    month = mar,
    year = "2024",
    address = "St. Julian{'}s, Malta",
    publisher = "Association for Computational Linguistics",
    url = "https://aclanthology.org/2024.findings-eacl.64/",
    pages = "946--964"
}

@inproceedings{shen-etal-2021-corpus,
    title = "Corpus-based Open-Domain Event Type Induction",
    author = "Shen, Jiaming  and
      Zhang, Yunyi  and
      Ji, Heng  and
      Han, Jiawei",
    editor = "Moens, Marie-Francine  and
      Huang, Xuanjing  and
      Specia, Lucia  and
      Yih, Scott Wen-tau",
    booktitle = "Proceedings of the 2021 Conference on Empirical Methods in Natural Language Processing",
    month = nov,
    year = "2021",
    address = "Online and Punta Cana, Dominican Republic",
    publisher = "Association for Computational Linguistics",
    url = "https://aclanthology.org/2021.emnlp-main.441/",
    doi = "10.18653/v1/2021.emnlp-main.441",
    pages = "5427--5440"
}

@inproceedings{edwards-ji-2023-semi,
    title = "Semi-supervised New Event Type Induction and Description via Contrastive Loss-Enforced Batch Attention",
    author = "Edwards, Carl  and
      Ji, Heng",
    editor = "Vlachos, Andreas  and
      Augenstein, Isabelle",
    booktitle = "Proceedings of the 17th Conference of the European Chapter of the Association for Computational Linguistics",
    month = may,
    year = "2023",
    address = "Dubrovnik, Croatia",
    publisher = "Association for Computational Linguistics",
    url = "https://aclanthology.org/2023.eacl-main.275/",
    doi = "10.18653/v1/2023.eacl-main.275",
    pages = "3805--3827"
}

@inproceedings{li-etal-2021-document,
    title = "Document-Level Event Argument Extraction by Conditional Generation",
    author = "Li, Sha  and
      Ji, Heng  and
      Han, Jiawei",
    editor = "Toutanova, Kristina  and
      Rumshisky, Anna  and
      Zettlemoyer, Luke  and
      Hakkani-Tur, Dilek  and
      Beltagy, Iz  and
      Bethard, Steven  and
      Cotterell, Ryan  and
      Chakraborty, Tanmoy  and
      Zhou, Yichao",
    booktitle = "Proceedings of the 2021 Conference of the North American Chapter of the Association for Computational Linguistics: Human Language Technologies",
    month = jun,
    year = "2021",
    address = "Online",
    publisher = "Association for Computational Linguistics",
    url = "https://aclanthology.org/2021.naacl-main.69/",
    doi = "10.18653/v1/2021.naacl-main.69",
    pages = "894--908"
}

@article{Raffel-etal-2020-exploring,
author = {Raffel, Colin and Shazeer, Noam and Roberts, Adam and Lee, Katherine and Narang, Sharan and Matena, Michael and Zhou, Yanqi and Li, Wei and Liu, Peter J.},
title = {Exploring the limits of transfer learning with a unified text-to-text transformer},
year = {2020},
issue_date = {January 2020},
publisher = {JMLR.org},
volume = {21},
number = {1},
issn = {1532-4435},
journal = {J. Mach. Learn. Res.},
month = jan,
articleno = {140},
numpages = {67}
}

@inproceedings{devlin-etal-2019-bert,
    title = "{BERT}: Pre-training of Deep Bidirectional Transformers for Language Understanding",
    author = "Devlin, Jacob  and
      Chang, Ming-Wei  and
      Lee, Kenton  and
      Toutanova, Kristina",
    editor = "Burstein, Jill  and
      Doran, Christy  and
      Solorio, Thamar",
    booktitle = "Proceedings of the 2019 Conference of the North {A}merican Chapter of the Association for Computational Linguistics: Human Language Technologies, Volume 1 (Long and Short Papers)",
    month = jun,
    year = "2019",
    address = "Minneapolis, Minnesota",
    publisher = "Association for Computational Linguistics",
    url = "https://aclanthology.org/N19-1423/",
    doi = "10.18653/v1/N19-1423",
    pages = "4171--4186"
}

@INPROCEEDINGS{liu-etal-event-recommandation,
  author={Liu, Chun-Yi and Zhou, Chuan and Wu, Jia and Xie, Hongtao and Hu, Yue and Guo, Li},
  booktitle={2017 International Joint Conference on Neural Networks (IJCNN)}, 
  title={CPMF: A collective pairwise matrix factorization model for upcoming event recommendation}, 
  year={2017},
  volume={},
  number={},
  pages={1532-1539},
  doi={10.1109/IJCNN.2017.7966033}}

@INPROCEEDINGS{wu-etal-event-graph,
  author={Wu, Xindong and Wu, Jia and Fu, Xiaoyi and Li, Jiachen and Zhou, Peng and Jiang, Xu},
  booktitle={2019 IEEE International Conference on Data Mining (ICDM)}, 
  title={Automatic Knowledge Graph Construction: A Report on the 2019 ICDM/ICBK Contest}, 
  year={2019},
  volume={},
  number={},
  pages={1540-1545},
  doi={10.1109/ICDM.2019.00204}}

@article{goran-etal-ir,
author = {Glava\v{s}, Goran and \v{S}najder, Jan},
title = {Event graphs for information retrieval and multi-document summarization},
year = {2014},
issue_date = {November, 2014},
publisher = {Pergamon Press, Inc.},
address = {USA},
volume = {41},
number = {15},
issn = {0957-4174},
url = {https://doi.org/10.1016/j.eswa.2014.04.004},
doi = {10.1016/j.eswa.2014.04.004},
journal = {Expert Syst. Appl.},
month = nov,
pages = {6904–6916},
numpages = {13}
}

@article{van2014tsne,
  author  = {Laurens van der Maaten},
  title   = {Accelerating t-SNE using Tree-Based Algorithms},
  journal = {Journal of Machine Learning Research},
  year    = {2014},
  volume  = {15},
  number  = {93},
  pages   = {3221--3245},
  url     = {http://jmlr.org/papers/v15/vandermaaten14a.html}
}

@inproceedings{2016evalonto,
author = {Dellschaft, Klaas and Staab, Steffen},
title = {On how to perform a gold standard based evaluation of ontology learning},
year = {2006},
isbn = {3540490299},
publisher = {Springer-Verlag},
address = {Berlin, Heidelberg},
url = {https://doi.org/10.1007/11926078_17},
doi = {10.1007/11926078_17},
booktitle = {Proceedings of the 5th International Conference on The Semantic Web},
pages = {228–241},
numpages = {14},
location = {Athens, GA},
series = {ISWC'06}
}

@inproceedings{lin-2004-rouge,
    title = "{ROUGE}: A Package for Automatic Evaluation of Summaries",
    author = "Lin, Chin-Yew",
    booktitle = "Text Summarization Branches Out",
    month = jul,
    year = "2004",
    address = "Barcelona, Spain",
    publisher = "Association for Computational Linguistics",
    url = "https://aclanthology.org/W04-1013/",
    pages = "74--81"
}

@inproceedings{bert-score,
  title={BERTScore: Evaluating Text Generation with BERT},
  author={Tianyi Zhang* and Varsha Kishore* and Felix Wu* and Kilian Q. Weinberger and Yoav Artzi},
  booktitle={International Conference on Learning Representations},
  year={2020},
  url={https://openreview.net/forum?id=SkeHuCVFDr}
}

@article{Yang2025Qwen3,
  title={Qwen3 Technical Report},
  author={Yang, An and Li, Anfeng and Yang, Baosong and Zhang, Beichen and Hui, Binyuan and others},
  journal={arXiv preprint arXiv:2505.09388},
  year={2025},
  note={Presents the Qwen3 series, including dense and MoE models, hybrid thinking mode, and the thinking budget mechanism}
}

@inproceedings{zeng-etal-2025-codetaxo,
    title = "{C}ode{T}axo: Enhancing Taxonomy Expansion with Limited Examples via Code Language Prompts",
    author = "Zeng, Qingkai  and
      Bai, Yuyang  and
      Tan, Zhaoxuan  and
      Wu, Zhenyu  and
      Feng, Shangbin  and
      Jiang, Meng",
    editor = "Che, Wanxiang  and
      Nabende, Joyce  and
      Shutova, Ekaterina  and
      Pilehvar, Mohammad Taher",
    booktitle = "Findings of the Association for Computational Linguistics: ACL 2025",
    month = jul,
    year = "2025",
    address = "Vienna, Austria",
    publisher = "Association for Computational Linguistics",
    url = "https://aclanthology.org/2025.findings-acl.214/",
    doi = "10.18653/v1/2025.findings-acl.214",
    pages = "4131--4144",
    ISBN = "979-8-89176-256-5"
}

@inproceedings{yu-etal-2020-steam,
author = {Yu, Yue and Li, Yinghao and Shen, Jiaming and Feng, Hao and Sun, Jimeng and Zhang, Chao},
title = {STEAM: Self-Supervised Taxonomy Expansion with Mini-Paths},
year = {2020},
isbn = {9781450379984},
publisher = {Association for Computing Machinery},
address = {New York, NY, USA},
url = {https://doi.org/10.1145/3394486.3403145},
doi = {10.1145/3394486.3403145},
booktitle = {Proceedings of the 26th ACM SIGKDD International Conference on Knowledge Discovery \& Data Mining},
pages = {1026–1035},
numpages = {10},
location = {Virtual Event, CA, USA},
series = {KDD '20}
}

@inproceedings{shen-etal-2020-taxoexpan,
author = {Shen, Jiaming and Shen, Zhihong and Xiong, Chenyan and Wang, Chi and Wang, Kuansan and Han, Jiawei},
title = {TaxoExpan: Self-supervised Taxonomy Expansion with Position-Enhanced Graph Neural Network},
year = {2020},
isbn = {9781450370233},
publisher = {Association for Computing Machinery},
address = {New York, NY, USA},
url = {https://doi.org/10.1145/3366423.3380132},
doi = {10.1145/3366423.3380132},
booktitle = {Proceedings of The Web Conference 2020},
pages = {486–497},
numpages = {12},
location = {Taipei, Taiwan},
series = {WWW '20}
}

@inproceedings{manzoor-etal-2020-expanding,
author = {Manzoor, Emaad and Li, Rui and Shrouty, Dhananjay and Leskovec, Jure},
title = {Expanding Taxonomies with Implicit Edge Semantics},
year = {2020},
isbn = {9781450370233},
publisher = {Association for Computing Machinery},
address = {New York, NY, USA},
url = {https://doi.org/10.1145/3366423.3380271},
doi = {10.1145/3366423.3380271},
booktitle = {Proceedings of The Web Conference 2020},
pages = {2044–2054},
numpages = {11},
location = {Taipei, Taiwan},
series = {WWW '20}
}

@inproceedings{doddington-etal-2004-automatic,
    title = "The Automatic Content Extraction ({ACE}) Program {--} Tasks, Data, and Evaluation",
    author = "Doddington, George  and
      Mitchell, Alexis  and
      Przybocki, Mark  and
      Ramshaw, Lance  and
      Strassel, Stephanie  and
      Weischedel, Ralph",
    editor = "Lino, Maria Teresa  and
      Xavier, Maria Francisca  and
      Ferreira, F{\'a}tima  and
      Costa, Rute  and
      Silva, Raquel",
    booktitle = "Proceedings of the Fourth International Conference on Language Resources and Evaluation ({LREC}{'}04)",
    month = may,
    year = "2004",
    address = "Lisbon, Portugal",
    publisher = "European Language Resources Association (ELRA)",
    url = "https://aclanthology.org/L04-1011/"
}

@inproceedings{song-etal-2015-light,
    title = "From Light to Rich {ERE}: Annotation of Entities, Relations, and Events",
    author = "Song, Zhiyi  and
      Bies, Ann  and
      Strassel, Stephanie  and
      Riese, Tom  and
      Mott, Justin  and
      Ellis, Joe  and
      Wright, Jonathan  and
      Kulick, Seth  and
      Ryant, Neville  and
      Ma, Xiaoyi",
    editor = "Hovy, Eduard  and
      Mitamura, Teruko  and
      Palmer, Martha",
    booktitle = "Proceedings of the 3rd Workshop on {EVENTS}: Definition, Detection, Coreference, and Representation",
    month = jun,
    year = "2015",
    address = "Denver, Colorado",
    publisher = "Association for Computational Linguistics",
    url = "https://aclanthology.org/W15-0812/",
    doi = "10.3115/v1/W15-0812",
    pages = "89--98"
}

@inproceedings{wang-etal-2020-maven,
    title = "{MAVEN}: {A} {M}assive {G}eneral {D}omain {E}vent {D}etection {D}ataset",
    author = "Wang, Xiaozhi  and
      Wang, Ziqi  and
      Han, Xu  and
      Jiang, Wangyi  and
      Han, Rong  and
      Liu, Zhiyuan  and
      Li, Juanzi  and
      Li, Peng  and
      Lin, Yankai  and
      Zhou, Jie",
    editor = "Webber, Bonnie  and
      Cohn, Trevor  and
      He, Yulan  and
      Liu, Yang",
    booktitle = "Proceedings of the 2020 Conference on Empirical Methods in Natural Language Processing (EMNLP)",
    month = nov,
    year = "2020",
    address = "Online",
    publisher = "Association for Computational Linguistics",
    url = "https://aclanthology.org/2020.emnlp-main.129/",
    doi = "10.18653/v1/2020.emnlp-main.129",
    pages = "1652--1671"
}

@article{hubert1985comparing,
  title={Comparing partitions},
  author={Hubert, Lawrence and Arabie, Phipps},
  journal={Journal of classification},
  volume={2},
  number={1},
  pages={193--218},
  year={1985},
  publisher={Springer}
}

@inproceedings{bagga-baldwin-1998-entity,
    title = "Entity-Based Cross-Document Coreferencing Using the Vector Space Model",
    author = "Bagga, Amit  and
      Baldwin, Breck",
    booktitle = "{COLING} 1998 Volume 1: The 17th International Conference on Computational Linguistics",
    year = "1998",
    url = "https://aclanthology.org/C98-1012/"
}

@inproceedings{huang-ji-2020-semi,
    title = "Semi-supervised New Event Type Induction and Event Detection",
    author = "Huang, Lifu  and
      Ji, Heng",
    editor = "Webber, Bonnie  and
      Cohn, Trevor  and
      He, Yulan  and
      Liu, Yang",
    booktitle = "Proceedings of the 2020 Conference on Empirical Methods in Natural Language Processing (EMNLP)",
    month = nov,
    year = "2020",
    address = "Online",
    publisher = "Association for Computational Linguistics",
    url = "https://aclanthology.org/2020.emnlp-main.53/",
    doi = "10.18653/v1/2020.emnlp-main.53",
    pages = "718--724"
}

@article{crouse2016implementing,
  title={On implementing 2D rectangular assignment algorithms},
  author={Crouse, David F},
  journal={IEEE Transactions on Aerospace and Electronic Systems},
  volume={52},
  number={4},
  pages={1679--1696},
  year={2016},
  publisher={IEEE}
}

\appendix

\section{Dataset Statistics}
\label{app:datasets}

Table~\ref{tab:eoc_dataset_statistics} reports the split of known and unknown event types used in our experiments.
Following HALTON~\cite{cao-etal-2023-event}, we simulate the realistic ontology expansion scenario where only frequent event types are covered by the initial ontology, while less frequent or newly emerging types must be discovered from unlabeled data.
This setting evaluates whether a model can expand an existing ontology beyond its predefined type inventory rather than merely classify instances into known categories.

\begin{table}[!htbp]
\centering
\small
\caption{Dataset statistics for event ontology expansion.}
\resizebox{\linewidth}{!}{%
\begin{tabular}{lcccc}
        \toprule
        \textbf{Dataset} & \textbf{\#Known} & \textbf{\#Unknown} & \textbf{\#Total} & \textbf{Unk. Ratio (\%)} \\
        \midrule
        ACE   & 10 & 23 & 33 & 69.7 \\
        ERE   & 10 & 28 & 38 & 73.7 \\
        MAVEN & 20 & 40 & 60 & 66.7 \\
        \bottomrule
    \end{tabular}
}
\label{tab:eoc_dataset_statistics}
\end{table}

The unknown-type ratios are high across all three datasets, ranging from 66.7\% to 73.7\%.
Therefore, the evaluation places substantial emphasis on discovering and organizing unseen event types.
MAVEN contains the largest number of total and unknown types, making it a more challenging benchmark for both clustering and hierarchy expansion because the model must distinguish a larger set of fine-grained event semantics.

\section{Baseline Details}
\label{app:baselines}

We include baselines that cover the three major abilities required by event ontology expansion: discovering unknown types, attaching discovered types to an ontology, and producing human-readable type names.
The goal is not only to compare against methods with similar architectures, but also to test whether conceptualization brings consistent benefits over representative alternatives at each stage.

\paragraph{Event clustering baselines.}
SS-VQ-VAE~\cite{huang-ji-2020-semi} is a semi-supervised event type induction method based on VQ-VAE.
ETYPECLUS~\cite{shen-etal-2021-corpus} clusters events with salient predicate-argument features in a latent spherical space.
TABS~\cite{li-etal-2021-document} injects abstraction-based semantic representations.
HALTON~\cite{cao-etal-2023-event} adopts neighborhood contrastive learning with K-Means pseudo labels.
These baselines allow us to compare ConceptE with semi-supervised clustering, unsupervised clustering, abstraction-based representations, and the strongest directly related event ontology expansion method.

\paragraph{Hierarchy expansion baselines.}
Clustering baselines + GE combine clustering-based discovery with greedy expansion.
Type\_Similarity links averaged cluster representations to encoded known type names.
LLMs\_Prompt directly prompts an LLM to select a parent type.
HALTON performs hierarchy-aware representation learning with path-based supervision and greedy top-down inference.
This group tests whether explicit hierarchy modeling is needed beyond flat clustering, and whether learned parent-child linking is more reliable than direct embedding similarity or direct LLM prompting.

\paragraph{Naming baselines.}
TABS uses abstraction-based generation; T5\_Template~\cite{Raffel-etal-2020-exploring} fills predefined templates; Trigger\_Sel selects a trigger word from the cluster; HALTON prompts an LLM with a cluster-center instance and predicted path; Top1\_Concept selects the most frequent concept name within a cluster.
The naming baselines represent template-based generation, trigger-based selection, LLM prompting with instance-level information, and a simple concept-frequency heuristic.
Comparing against Top1\_Concept is particularly useful because it isolates whether using only the most frequent generated concept is sufficient.
The results show that conditioning on multiple representative concepts and the predicted ontology path is more effective.

\section{Type Naming Details}
\label{app:type_naming_details}

The type naming stage is formulated as ontology-aware generation.
Given a discovered cluster, we provide the LLM with representative concepts from the cluster and its predicted parent path.
The representative concepts summarize the cluster semantics, while the parent path exposes the naming convention and semantic scope expected by the ontology.
The prompt instructs the LLM to generate a concise event type name rather than a free-form description.
This design is important because event type names are not arbitrary summaries: they must match the granularity of the existing ontology and remain compatible with neighboring nodes in the hierarchy.
For example, a name that is too broad may overlap with its parent type, while a name that is too specific may describe only a subset of instances in the cluster.

\begin{figure}[!htbp]
  \centering
  \includegraphics[width=1.0\linewidth]{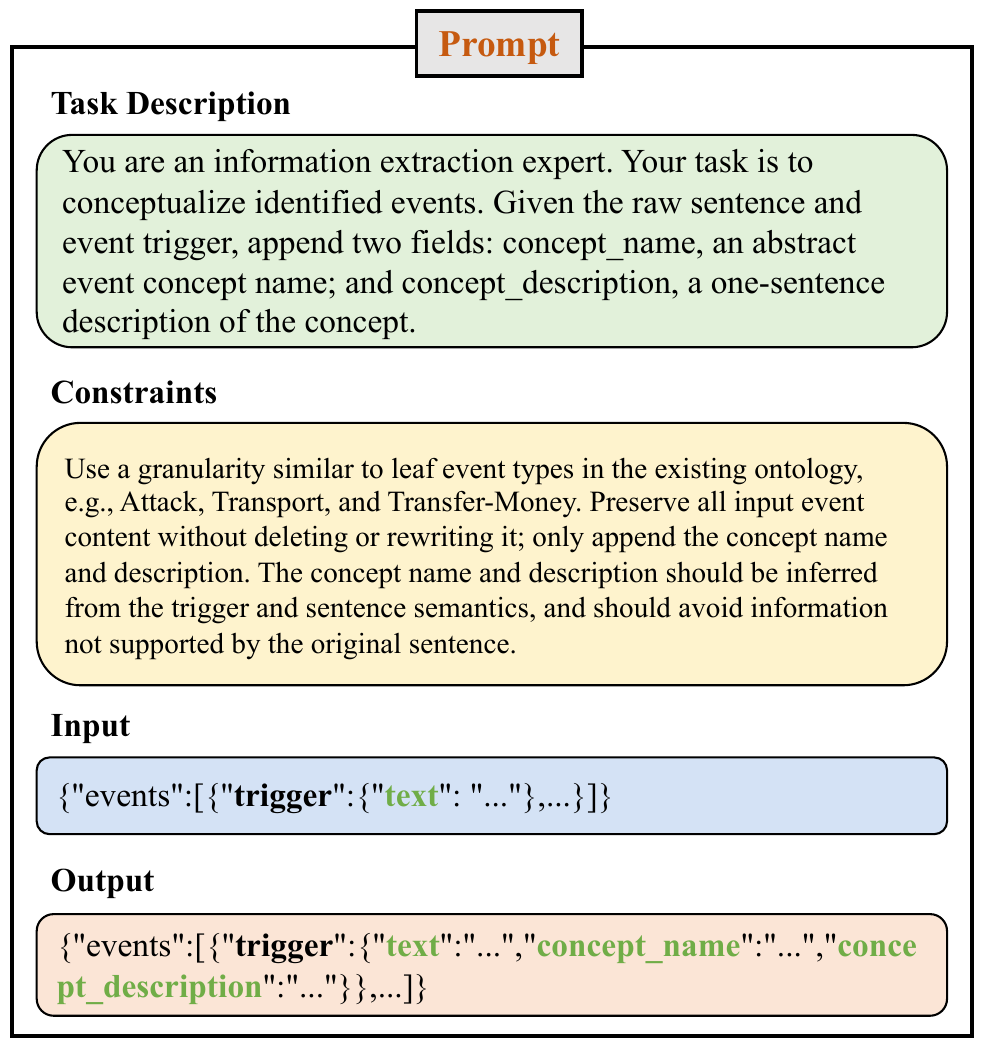}
  \caption{Illustration of the conceptualization prompt. The prompt takes the input sentence and the identified trigger, and instructs the LLM to generate concept names and descriptions for the trigger.}
  \label{fig:conceptualization_prompt}
\end{figure}

\begin{figure}[!htbp]
\centering
\includegraphics[width=1.0\linewidth]{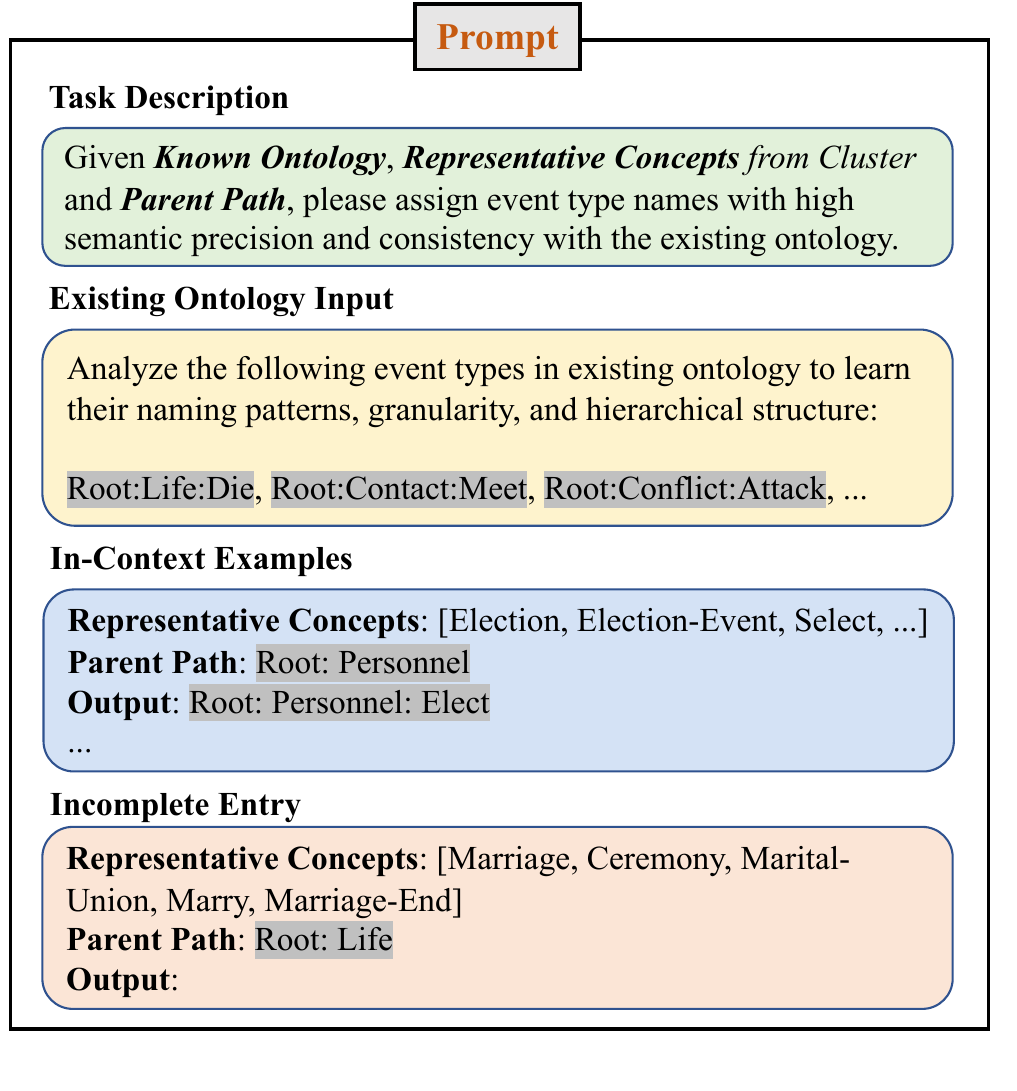}
\caption{An example prompt for ontology-aware event type naming.}
\label{fig:naming_prompt}
\end{figure}

Figure~\ref{fig:naming_prompt} shows the full prompt used for type naming.
For completeness, Figure~\ref{fig:conceptualization_prompt} also shows the conceptualization prompt that produces the concept names used by the naming module.
The prompt examples are moved to the appendix to keep the main text focused on the modeling idea.
They also provide reproducibility details for how concepts and names are elicited from the LLM.

\section{Conceptualization Details}
\label{app:concept_details}
To reduce generation noise and hallucination, the conceptualization prompt imposes three constraints.
First, concept names are required to be concise schema-level categories with granularity comparable to event ontology leaf types.
Second, the input sentence and trigger must be preserved exactly, and the model should only output the concept name and concept description.
Third, the concept name and description must be grounded in the sentence containing the trigger rather than unsupported external information.

Concept granularity is important for ontology expansion.
If concepts are too coarse, different unknown event types become hard to separate; if they are too fine-grained, the representation again becomes sensitive to entities, time expressions, or incidental details.
We therefore use concepts at the level of candidate event-type phrases: more abstract than a surface trigger, but not tied to a specific instance.

ConceptE uses conceptualization as an intermediate semantic abstraction rather than as a replacement for the original event trigger.
The original sentence and trigger preserve trigger-level information, while the generated concept exposes the concept-level semantics that are often implicit in the sentence.
Jointly encoding these signals helps the model retain discriminative lexical cues while reducing sensitivity to incidental contextual variation.

In practice, the generated concept description is useful when the concept name alone is ambiguous.
For example, a short concept such as \emph{Transfer} may correspond to different event semantics depending on whether the context involves people, ownership, money, or organizational control.
The description provides a grounded semantic explanation, allowing the encoder to distinguish such cases more reliably.

\section{Feature Analysis}
\label{app:feature_analysis}

The parent-child linker predicts whether a newly discovered type should be attached under a candidate ontology node.
Since this relation is directional, a pure similarity score is insufficient.
A child type and a parent type may be semantically related, but the direction of subsumption still needs to be modeled.
We therefore compare several interaction feature combinations in Table~\ref{tab:feature_analysis}.

\begin{table}[!htbp]
\centering
\small
\caption{Comparison of different interaction feature combinations in the parent-child linker on the ACE dataset.}
\label{tab:feature_analysis} 
\resizebox{\linewidth}{!}{%
\begin{tabular}{l|c|c}
    \toprule
    \multirow{2}{*}{Methods} & \multicolumn{2}{c}{Taxo\_F1 (\%)} \\ 
                            & Predicted Cluster & Golden Cluster \\ 
    \midrule
    $[\mathbf{u};\mathbf{v};\mathbf{u}\odot\mathbf{v};|\mathbf{u}-\mathbf{v}|]$             & \textbf{43.48}  & \textbf{51.16}          \\ 
    $[\mathbf{u};\mathbf{v};\mathbf{u}\odot\mathbf{v}]$                & 41.62 & 48.32          \\ 
    $[\mathbf{u};\mathbf{v};|\mathbf{u}-\mathbf{v}|]$              & 38.36 & 44.82           \\ 
    $[\mathbf{u};\mathbf{v}]$          & 35.56 &  40.21        \\ 
    \bottomrule
\end{tabular}
}
\end{table}

The full feature set performs best in both predicted-cluster and golden-cluster settings.
Concatenation preserves the identity of the child and parent representations, element-wise product captures compatibility, and absolute difference captures mismatch between the two nodes.
Removing either product or difference features degrades performance, indicating that parent-child linking benefits from both similarity-oriented and distance-oriented interactions.
Using only concatenation performs worst, which suggests that the linker needs explicit cross-node interaction features rather than relying solely on an MLP to infer all pairwise relations from raw embeddings.

\section{Implementation Details}
\label{app:implementation}

This section provides additional training and inference details for reproducibility.
All modules are trained under the same dataset splits as HALTON, and the number of unknown clusters is set to the number of unknown event types in each benchmark.
Unless otherwise stated, hyperparameters are selected on the development set and then kept fixed for the reported test results.

\paragraph{Event Clustering.}
For event clustering, we use BERT-base-uncased~\cite{devlin-etal-2019-bert} as the encoder and use the 12th-layer hidden states for representation extraction.
The maximum input length is 240, the hidden dimension is 512, and the clustering space dimension is 256.
Trigger spans and concept spans are pooled with max pooling.
We train the clustering module with Adam using a learning rate of $1\times10^{-4}$, batch size 64, and 60 epochs.
The supervised contrastive temperature is 0.07, and the center loss weight is 0.005.
For unlabeled neighborhood mining, $K=2$ for ACE and ERE and $K=15$ for MAVEN.
The $k$-means pseudo labels are refreshed once per epoch with the number of clusters set to the number of unknown event types.
Event conceptualization is performed using Qwen3-32B~\cite{Yang2025Qwen3}.
During concept generation, we set temperature to 0.7, $top_p$ to 0.8, $top_k$ to 20, the maximum token budget to 8192, and disable the model's thinking mode.
The memory bank for neighborhood mining is updated with the latest instance representations at each epoch.
This allows nearest-neighbor positives for unlabeled data to reflect the current representation space rather than a fixed initialization.
We use a larger neighborhood size for MAVEN because it contains more unknown event types and more diverse event mentions.

\paragraph{Hierarchy Expansion.}
For hierarchy expansion, we reuse the BERT encoder trained in the clustering stage and freeze all its parameters.
We then train only the hierarchy modules, including the recursive concept composition module and the directional parent-child linker.
The hierarchy stage uses Adam with a learning rate of $1\times10^{-3}$, batch size 256, and 100 epochs.
The cosine alignment loss for the composition module is weighted by 0.1.
For hierarchy prediction, we use weighted binary cross-entropy with positive and negative edge weights both set to 2.0.
For each child node, up to 10 negative parent nodes are sampled from non-ancestor nodes; when available, same-level non-ancestor nodes are preferred as hard negatives.
We use the trained clustering checkpoint to obtain cluster representations for the predicted-cluster setting and gold labels for the golden-cluster setting.
The predicted-cluster setting evaluates the full pipeline, where hierarchy expansion must handle errors propagated from event clustering.
The golden-cluster setting isolates the hierarchy module by removing clustering errors, allowing us to test whether the ontology modeling component itself is effective.

\paragraph{Type Naming.}
For type naming, we employ Qwen3-32B with the same decoding configuration as in conceptualization.
The naming prompt takes the existing ontology context, high-frequency concept names from the target cluster, and the predicted parent path as input.
We use high-frequency concepts because they are more likely to represent the shared semantics of the cluster rather than noisy outlier mentions.
The predicted parent path is included to encourage names that are consistent with the semantic scope and naming style of the target ontology region.

\section{Event Clustering Visualization}
\label{app:visualization}

\begin{figure*}[!htbp] 
\centering 
\includegraphics[width=0.92\textwidth]{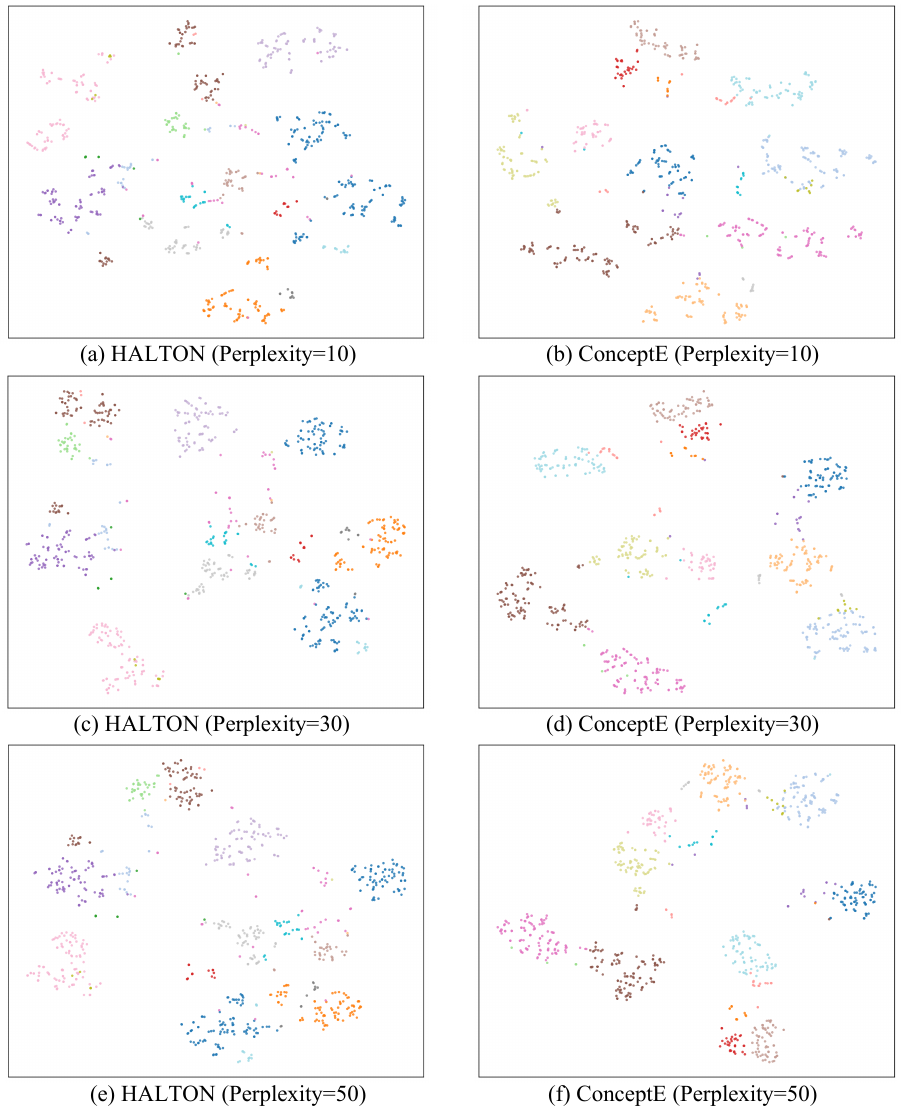} 
\caption{Visualization of event clustering features under different t-SNE perplexity values.}
\label{fig:clustering_vis} 
\end{figure*}

To provide an intuitive comparison of clustering quality and examine whether the visualization trend is stable across t-SNE hyperparameters, we project event representations on the ERE dataset into two dimensions using t-SNE~\cite{van2014tsne}.
Because t-SNE is sensitive to perplexity, we set the perplexity to 10, 30, and 50, and visualize the event representations of HALTON and ConceptE under each setting.
Figure~\ref{fig:clustering_vis} shows six results: (a), (c), and (e) correspond to HALTON under the three perplexity values, while (b), (d), and (f) correspond to ConceptE under the same settings.
Across different perplexity values, HALTON shows less compact clusters and less clear boundaries, whereas ConceptE produces more stable local groups and clearer separation.
This qualitative trend supports the quantitative clustering results in Table~\ref{tab:cluster_performance}.

We emphasize that t-SNE visualizations should be interpreted as qualitative evidence only.
Distances and cluster shapes in two-dimensional projections can be affected by random initialization, perplexity, and optimization dynamics.
For this reason, our main conclusions rely on quantitative clustering metrics, while the visualization is used to illustrate that the improvement brought by conceptualization is also visible under multiple projection settings.

\end{document}